\documentclass{article}
\usepackage{spconf,amsmath,graphicx,multirow}
\usepackage{enumitem}
\title{VU-BERT: A Unified framework for Visual Dialog}
\name{Tong Ye$^{1,2}$,  Shijing Si$^{1}$, Jianzong Wang$^{1,*}$\thanks{$^*$Corresponding author: Jianzong Wang, jzwang@188.com.}, Rui Wang$^{3}$, Ning Cheng$^{1}$, Jing Xiao$^{1}$}
\address{$^{1}$Ping An Technology (Shenzhen) Co., Ltd.\\
$^{2}$University of Science and Technology of China\\
$^{3}$Duke University}
\begin{document}
%\ninept
%
\maketitle
\begin{abstract}
The visual dialog task attempts to train an agent to answer multi-turn questions given an image, which requires the deep understanding of interactions between the image and dialog history. Existing researches tend to employ the modality-specific modules to model the interactions, which might be troublesome to use. To fill in this gap, we propose a unified framework for image-text joint embedding, named VU-BERT, and apply patch projection to obtain vision embedding firstly in visual dialog tasks to simplify the model. The model is trained over two tasks: masked language modeling and next utterance retrieval. These tasks help in learning visual concepts, utterances dependence, and the relationships between these two modalities. Finally, our VU-BERT achieves competitive performance (0.7287 NDCG scores) on VisDial v1.0 Datasets. 
%Lastly, we demonstrate detailed ablation studies to prove that VU-BERT makes full use of multi-modal information to improve performance.
\end{abstract}
\begin{keywords}
Multi-Modal, Visual Dialog, Patch Embedding, Transformer 
\end{keywords}
\section{Introduction}
\label{sec:intro}

Interactions between humans and the world  generally involve joint perception of information in miscellaneous modalities, \emph{e.g.}, image, text and audio. On the contrary, most of the well-known machine learning models are modality-specific like BERT \cite{bert,si2020students} or ResNet \cite{resnet}, agnostic to the information from interweaving of different modalities. So motivated, multi-modal tasks have recently gained increasing popularity, especially in the fields of vision and language.
At present, popular visual and language tasks include Visual Caption (VC) \cite{kim2021vilt,2}, Visual Grounding \cite{3,4}, Visual Question Answering (VQA) \cite{kim2021vilt,4,bai2021decomvqanet} and Visual Dialog (VD) \cite{schwartz2019factor,9,vd}. VQA attempts to predict a correct answer to questions given some background texts and images. The task of VD is firstly proposed by \cite{11}, where a dialog agent needs to answer a series of questions grounded by an image. Compared to VQA, VD is more challenging as the model needs not only to present high-level semantic understanding according to image and dialog history, but also to process and reason through the longer contexts than VQA.

To address VD tasks, existing works generally encode the input contexts and images separately with modality-specific modules, then combine the information from the two sources with a variety of cross-modal attention mechanisms, as in DVAN \cite{guo2019dual}, DualVD \cite{jiang2020dualvd}, VisDial-BERT \cite{murahari2020large}, \emph{etc}. Different from them, our VU-BERT leverages a single-stream transformer-based model to learn interactions, which jointly encodes information from the images and text without modality-specific modules. Most closely to our VU-BERT is VD-BERT \cite{vd}, which also models interactions by a unified transformer. However, VD-BERT must concatenate every answer candidate with the input and go through a forward pass of the entire model, which requires high computing resources and cannot scale to a large number of candidates like ours. Besides, our VU-BERT adapts patch embedding firstly in VD tasks while all of prior models process images complicatedly and inefficiently by ResNet \cite{resnet} or Faster R-CNN \cite{ren2015faster}.
To explore the interactions between images and dialog history, our framework utilizes BERT \cite{bert} architecture, and trains the model over two objectives: masked language modeling and next utterance retrieval. Different from previous work, we adopt a tractable visual embedding introduced in ViT \cite{vit}, which yields significant reduction on latency time and parameter scale \cite{kim2021vilt}. The key contributions of our study can be summarized as follows:
\begin{itemize}
%[itemsep=5pt,topsep=0pt,parsep=0pt,partopsep=0pt,leftmargin=15pt]
\item We propose a unified framework VU-BERT for effective vision and dialog representations to predict the next utterance by selecting or generating responses. And we achieve competitive performance on visual dialog tasks.
\item  Different from existing works \cite{schwartz2019factor,9,vd}, we attempt to obtain visual embedding by patch projection to the VD tasks, which leads to a simple architecture without any region features or deep convolution operations.
\item We conduct extensive experiments to verify the effectiveness of our framework. According to the ablation studies, we found that different dialog turns affect the performance, which indicates that it is not suitable to apply models specific to VQA or VC tasks directly to VD tasks.
\end{itemize}

\begin{figure*}[ht]
	\centering
	\includegraphics[width=1\textwidth]{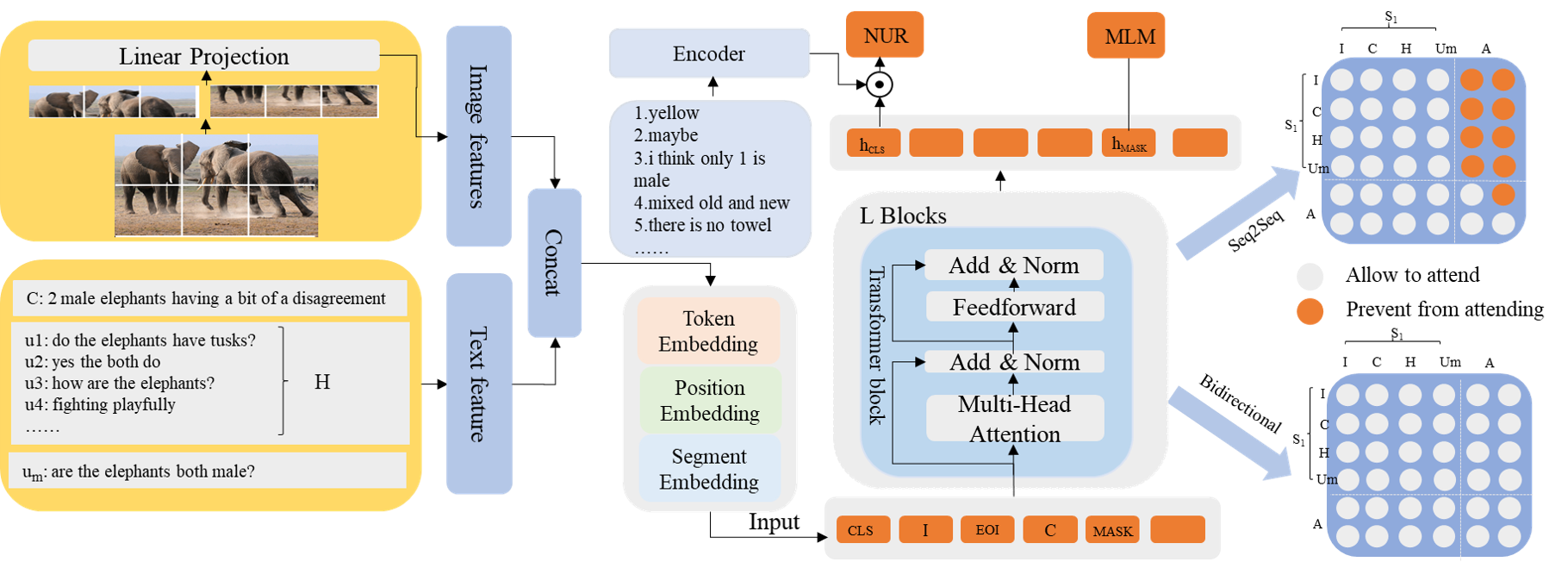}
	\caption{Overview of our VU-BERT. The image $I$ is split to patches and linearly projected to obtain patch embedding. The input consists of image and text embedding via summing up its position embedding and segment embedding. We keep the input as $(I,C,H,U_m)$ and the response candidates $A$ respectively passing through a transformer encoder.}
	\label{fig1}
\end{figure*}

\section{methodology}
\label{sec:model}
In this section, we discuss our visual dialog model VU-BERT, including two novel approaches aimed at capturing better representations of vision and utterances as shown in Fig \ref{fig1}. We first employ a unified vision-dialog Transformer to encode both the image and dialog history. Then, we adopt visually grounded masked language modeling for effective interactions between visual and textual features and next utterance retrieval for enhancing utterances dependence to train the model. Next, we describe the components of this model in detail.

\subsection{Embeddding}
\label{ssec:embedding}

\textbf{Dialog Embedding.}
Given a $m$-th turn dialog represented by a list of utterances $U=[u_1,\ldots, u_m]$ grounded in the image $I$, an utterance is first split into words $u_i=[w_1,\ldots,w_n]$ with length of $n$ by the same WordPiece tokenizer \cite{16}. We follow BERT and each sub-word is obtained via summing up its word embedding, position embedding and segment embedding.

\noindent \textbf{Vision Embedding.} Different from extracting object-level features as vision features \cite{schwartz2019factor,9,vd}, we use ViT\cite{vit} as a backbone network to process images, which is faster than object detectors. According to ViT, 
Patch Embedding will split the height and width of the input image $I\in R^{H\times W\times C}$  to $N=\frac{HW}{P^2}$ patches according P, and then flatten and reshape the patches to $v\in R^{N\times\left(P^2\times C\right)}$ through linear transformation.
%the input image $I\in R^{H\times W \times C}$ is split into a sequence of patches and flattened into $N=HW/P^2$ patches $v\in R^{N\times (P^2\times C)}$ where $C$ is the number of channels, $(H,W)$ is the input image resolution and $(P,P)$ is the patch resolution. 
%The patches are linearly projected to obtain patch embedding. 
Similar to word embedding in BERT, we add segment embedding and position embedding to patch embedding as visual embedding where all the patches share the same segment embedding.
\subsection{Network Architecture}
VU-BERT, which unifies the multi-model architecture into a single model, is described in Fig.1.  We prepare the input sequence as $x=\{[CLS]$, $v_1$, $\ldots$, $v_t$, $[EOI], C$, $[SEP]$, $u_1$, $[SEP],\ldots,u_m,[END]\}$. The special token $[CLS]$ and $[END]$ denote the beginning and the end of the sequence respectively and $[SEP]$ separates the different utterances. In addition, we add the token $[EOI]$ to indicate the end of image input. Moreover, we combine the input sequence with its position embedding and segment embedding. Then we encode the sequences into cross-modal representations $H=[h_{cls}, h_1,\ldots, h_{end}]$ using the framework of Transformer. %Assuming the features of the $l$-th layer is $H^l$ ,where the $H^0$ denote the input elements, the features of $(l+1)$-th layer is computed as follows:
%\begin{align}
%& Q^{l+1}=H^{l}W_Q^{l+1},K^{l+1}=H^{l}W_K^{l+1},V^{l+1}=H^{l}W_V^{l+1} \\
%& M=
%\begin{cases}
%	0& \text{allow to attend}\\
%	-\infty& \text{prevent from attending}
%\end{cases} \\
%& Att^{l+1}=softmax(\frac{Q^{l+1}K^{l+1}}{\sqrt{d_k}}+M)V^{l+1} \\
%& h^{l+1}=LayerNorm(H^l+Att^{l+1}) \\
%& \tilde{H}_{l+1}=W_2^{l+1}\cdot GELU(W_1^{l+1}h^{l+1}+b_1^{l+1})+b_2^{l+1}\\
%& H^{l+1}=LayerNorm(\tilde{H}_{l+1}+h^{l+1}) &
%\end{align}
According to the seq2seq self-attention masks \cite{vlp,unified}, we mask out all next-turn utterance tokens by setting $-\infty$ in $M$ to get auto-regressive attentions while keeping tokens in the image $I$, caption $C$, dialog history $H$ and current utterance $U_m$ visible, which can recursively generate an answer by predicting the masked tokens. Similarly, we train VU-BERT with the NUR loss using bidirectional self-attention masks to select a correct response from response candidates. 
\subsection{Training Objectives}
\label{sec:PRE-TRAINING Objectives}

To obtain a strong and general representation for the model, we train VU-BERT with the following two losses:
\begin{table*}[t] \centering
	\setlength{\tabcolsep}{4mm}{
		\caption{Performance of discriminative/generation models on VisDial v0.9 dataset. Higher is better for MRR and recall@k, while lower is better for Mean.}\label{tab:my-table1}
		\begin{tabular}{lccccc}
			\hline
			Model   & MRR $\uparrow$           & R@1 $\uparrow$         & R@5 $\uparrow$         & R@10 $\uparrow$        & Mean $\downarrow$       \\ \hline
			LF      & 0.5807/0.5199 & 43.82/41.83 & 74.68/61.78  & 84.07/67.59 & 5.78/17.07 \\
			HRE     & 0.5846/0.5237 & 44.67/42.29 & 74.50/62.18 & 84.22/67.92 & 5.72/17.07 \\
			DVAN    & 0.6675/0.5594 & 53.62/46.58 & 82.85/65.50 & 90.72/71.25 & 3.93/14.79 \\
			FGA     & 0.6525/0.4138 & 51.43/27.42 & 82.08/56.33 & 89.56/71.32 & 4.35/\textbf{9.10}   \\
			VD-BERT & \textbf{0.7004/0.5595} & \textbf{55.79/46.83} & \textbf{85.34/65.43} & \textbf{92.68/72.05} & \textbf{4.04}/13.18 \\ \hline
			VU-BERT & 0.6333/0.5403  &48.71/44.50 &81.03/62.60        & 89.10/71.70  &4.19/12.49         \\ \hline
	\end{tabular}}
	\vspace{-0.5cm}	
	
\end{table*}
\subsubsection{Masked Language Modeling (MLM)}
Similar to MLM in BERT, each token in utterance is randomly masked out with 15\% probability and replaced with a special token [MASK]. The model is trained to predict the masked words, based on the unmasked words and the visual features. The task drives the network to not only model the dependencies in sentence words, but also to align the visual and utterance contents. We denote the visual embedding as $v= {v_1,\ldots,v_t}$ and the mask indices as $k$, where $w_{\backslash k}$ represents as $w_{\backslash k}=\{w_1,\ldots,w_{k-1},[MASK],w_{k+1},\ldots,w_n\}$. The parameters are learned to minimize the cross-entroy loss computed using the predicted tokens and the original tokens. Then the MLM loss function is shown in Eq \eqref{mlm}:
\begin{align}
    L_{mlm}=-E_{(w,v)\sim D} \log P(w_k|w_{\backslash k},v) \label{mlm}
\end{align}

\subsubsection{Next Utterance Retrieval (NUR)}
Given the utterances $[u_1,\ldots, u_{m-1}]$, the aim of NUR is to select the correct next utterance $u_m$ from a set of $N$ candidate responses. For this task, we need to respectively encode entire dialog context and each candidate response by using BERT, which allows for the precomputation of all possible candidates and can scale to millions of candidates. A list of candidate responses can be created automatically, with the positive sample being the next utterance in the dialog history, and making the assumption that randomly sampled utterances from different dialogues are likely to be negatives. We calculate the probability of $u_i$ being the true next utterance and minimize the cross-entropy loss of the next utterance, which is analogous to language modeling. The NUR loss function is:
\begin{align}
	L_{nur} &= -\log p(u_i|h_{cls},C) \notag \\
	&=-\log\bigg(\frac{\exp (h_{cls}*c_i)}{\exp (h_{cls}*c_i)+\sum \exp (h_{cls}*c_i^\prime)}\bigg)
\end{align}

Finally, we define the total loss as $L=\alpha L_{mlm}+\beta L_{nur}$. It's worth noting that we use the NUR task for effective vision and dialog fusion rather than NSP or ITM tasks. Because these tasks are not only weaker than seq2seq or bidirectional LM but also computationally expensive \cite{vlp}. 

%\subsection{More Details}
%When training for the answer generation, we mask out all answer tokens by setting $-\infty$ in $M$ to get auto-regressive attentions while keep tokens in the caption, dialog history and current question visible. \par
%If training for the answer selection, we keep the input as $(I,C,H,Q)$ and the answer candidates $A$ respectively passing through an BERT encoder. Then we use the dot product to output the answer according to a softmax layer. It's worth noting that we use the NUR task rather than Next Sentence Prediction tasks or Image-Text Matches. Because these tasks are not only weaker than seq2seq LM but also computationally expensive\cite{vlp}, which is agree with RoBERTa. 
%According to the MLM and NUR, we define our total loss as $L= L_{mlm} + L_{nur} $

\section{EXPERIMENT}
\label{sec:experiment}

\textbf{Datasets.}
We train VU-BERT on the large visual dialog datasets VisDial v0.9 and v1.0. Specifically, the v1.0 contains 123,287 images for training, 2,064 images for validation and 8,000 images for test,  which is a re-organization of train and validation splits from v0.9 to form the new train. Each image for training and validation is associated with one caption and 10 question-answer pairs. For each question, it is paired with a list of 100 answer candidates, one of which is regarded as the correct answer. In addition, visdial v1.0 provides dense annotations that specify real-valued relevance scores for the 100 answer candidates.
% We only collect train and validation splits in each dataset to avoid seeing any test data in pre-training. 
%Note that we did not pretrain our model in large-scale vision-language datasets like Conxeptual Captions and Visual Question Answering as other pretrained models do.

\noindent
\textbf{Baseline.}
We compare VU-BERT against a variety of single and ensemble baselines in VD tasks, including FGA \cite{schwartz2019factor}, VD-BERT \cite{vd}, LF \cite{11}, HRE \cite{11}, DVAN \cite{guo2019dual}, DualVD \cite{jiang2020dualvd}, VisDial-BERT \cite{murahari2020large} and CAG \cite{guo2020iterative}, which yield prominent performance from the leadboard or publication.

\noindent
\textbf{Experimental Setup.}
%As our model is similar to add new inputs to capture vision features, 
We use $BERT_{BASE}$ as the backbone, which consists of 12 Transformer blocks, each with 12 attention heads and a hidden state dimension of 768. We set dropout probability to 0.1 and the learning rate with Adam optimizer to 3e-4 and use GELU as activation function. In our experiments, we split each $224\times 224$ image into a $7\times 7$ grid of image patches, where each patch is $32\times 32$. We first train our model for 20 epochs with a batchsize of 32 on a cluster of 8 Tesla V100 with 16G memory using MLM and NUR losses. \par
We evaluate VU-BERT on VisDial v0.9 and v1.0 by using the ranking metrics including \textbf{Recall@K} ($K\in \{1, 5, 10\})$: the percentage of the ground truth answer option in top-k ranked responses, \textbf{Mean}: the average rank of the ground truth answer option, and \textbf{MRR} (Mean Reciprocal Rank): reciprocal rank of the ground truth answer option. Due to the acquisition of dense annotation, we add the \textbf{NDCG} (Normalized Discounted Cumulative Gain), which penalizes the lower rank of answers with high relevance, to evaluate each answer candidate for v1.0.

% \begin{figure}[t]
% 	\centering
% 	\includegraphics[width=0.5\textwidth, height=0.4\linewidth]{D:/document/111.png}
% 	\caption{Ranksing score }
% 	\label{fig2}
% \end{figure}

\section{Results and Analyses}
\label{sec:pagestyle}

\begin{table}[t]
	\setlength{\tabcolsep}{0.4mm}{
	\caption{Performance of discriminative models on VisDial v1.0 dataset. Higher is better for MRR, recall@k and NDCG, while lower is better for Mean.}\label{tab:my-table2}   
		\begin{tabular}{lcccccc}
			\hline		
		Model   & MRR$\uparrow$   & R@1$\uparrow$   & R@5$\uparrow$   & R@10$\uparrow$  & Mean$\downarrow$ & NDCG$\uparrow$  \\ \hline
		NMN     & 0.5880 & 44.15 & 76.88 & 86.88 & 4.81 & 0.5810 \\
		GNN       & 0.6137 & 47.33 & 77.98 & 87.83 & 4.57 & 0.5282 \\
		DualVD     &0.6323  &49.25  &80.23  &89.70  &\textbf{4.11}  &0.5632  \\
		CAG    & 0.6349 & \textbf{49.85} & 80.63 & \textbf{90.15} & \textbf{4.11} & 0.5664 \\ 
		LF           & 0.5542 & 40.95 & 72.45 & 82.83 & 5.95 & 0.4531 \\
		FGA          & \textbf{0.6370} & 49.58 & \textbf{80.97} & 88.55 & 4.51 & 0.5210 \\
		VD-BERT      & 0.5117 & 38.90 & 62.82 & 77.98 & 6.69 & \textbf{0.7454} \\
		VisDial-BERT & 0.5074 & 37.95 & 64.13 & 80.00 & 6.53 & 0.7447 \\ \hline
		VU-BERT &0.4909   &33.60 &67.20        &81.60 &6.12 &0.7287       \\ 
		\hline
	\end{tabular}}	
\vspace{-0.3cm}
\end{table}

\begin{table}[]
	\setlength{\tabcolsep}{1mm}{
		\caption{The parameters for image extraction models}
	\begin{tabular}{llll}
		\hline
		Image model &
		VGG &
		Fast R-CNN &
		Patch \\ \hline
		Method &
		\begin{tabular}[c]{@{}l@{}}LF;HRE; \\NMN; GNN;\\ DVAN; FGA\end{tabular} &
		\begin{tabular}[c]{@{}l@{}}DualVD; CAG; \\ VD-BERT; \\VisDial-BERT\end{tabular} &
		VU-BERT \\ 
		\# Parameters &
		138.36M &
		41.76M &
		\textbf{0.59M} \\ \hline
\end{tabular}}
\vspace{-0.3cm}	
	\label{tab:my-table3}
\end{table}

%In this section, we compare our model results with previous best published results on Visdial v0.9 and v1.0. Then we conduct ablation studies to examine various aspects of our model. Lastly, we interpret how it attains the effective fusion of vision and dialog via attention visualization.

\subsection{Results}
\textbf{On VisDial v0.9.} We show both discriminative and generative results on v0.9 with the state-of-the-art baselines from previous published results and leaderboard in Table \ref{tab:my-table1}. VU-BERT yields comparable results with the state of the art in the discriminative and generative settings, for which the MRR is 0.6333 and 0.5403 respectively. This validates the effectiveness of our model using a unified Transformer encoder.

\noindent
\textbf{on VisDial v1.0.} As shown in Tabel \ref{tab:my-table2}, we compare our results on VisDial v1.0 datasets.
We can found that VU-BERT outperforms most of the model across various metrics. Compared to VD-BERT, VU-BERT results in 4\% improvement in R@5, 3\% improvement in R@10 and 4\% reduction in Mean while in 2\% reduction in NDCG and in 2\% reduction in MRR. This indicates that VU-BERT can achieve comparable performance even without deep convolution. 

Obviously, VU-BERT is a simpler unified model without any region features or deep convolution operations. We commission the unified transformer module to process image and text simultaneously in place of an additional image extractor like Fast R-CNN, which leads to significant time and parameter efficiency \cite{kim2021vilt}. We notice that VU-BERT encodes candidate sets independently rather than integrates each candidate response at the input layer. Therefore, VU-BERT loses the early interaction between the input and the candidates, which coincidentally agrees with the results of some metrics lower than VD-BERT. In spite of a slight decline on metrics, VU-BERT can easily extend the candidate sets to a large amount, which facilitates adaptation to other tasks.
 
\noindent
\textbf{Parameters.} In order to present the lightweight VU-BERT, we compare the Patch Projection with two image extraction models commonly used in baseline methods: VGG and Fast R-CNN in Table 3. We can observe that the Patch Projection is much lighter when capturing the image feature. Then we selected several structurally similar models, which deal with text features by transformers, to compare the number of parameters in Table 4. Obviously, VU-BERT is much lighter than existing transformer-based models.

\subsection{Ablation Study}
\begin{figure}[t]
 	\centering
	\includegraphics[height=3.7cm,width=0.5\textwidth]{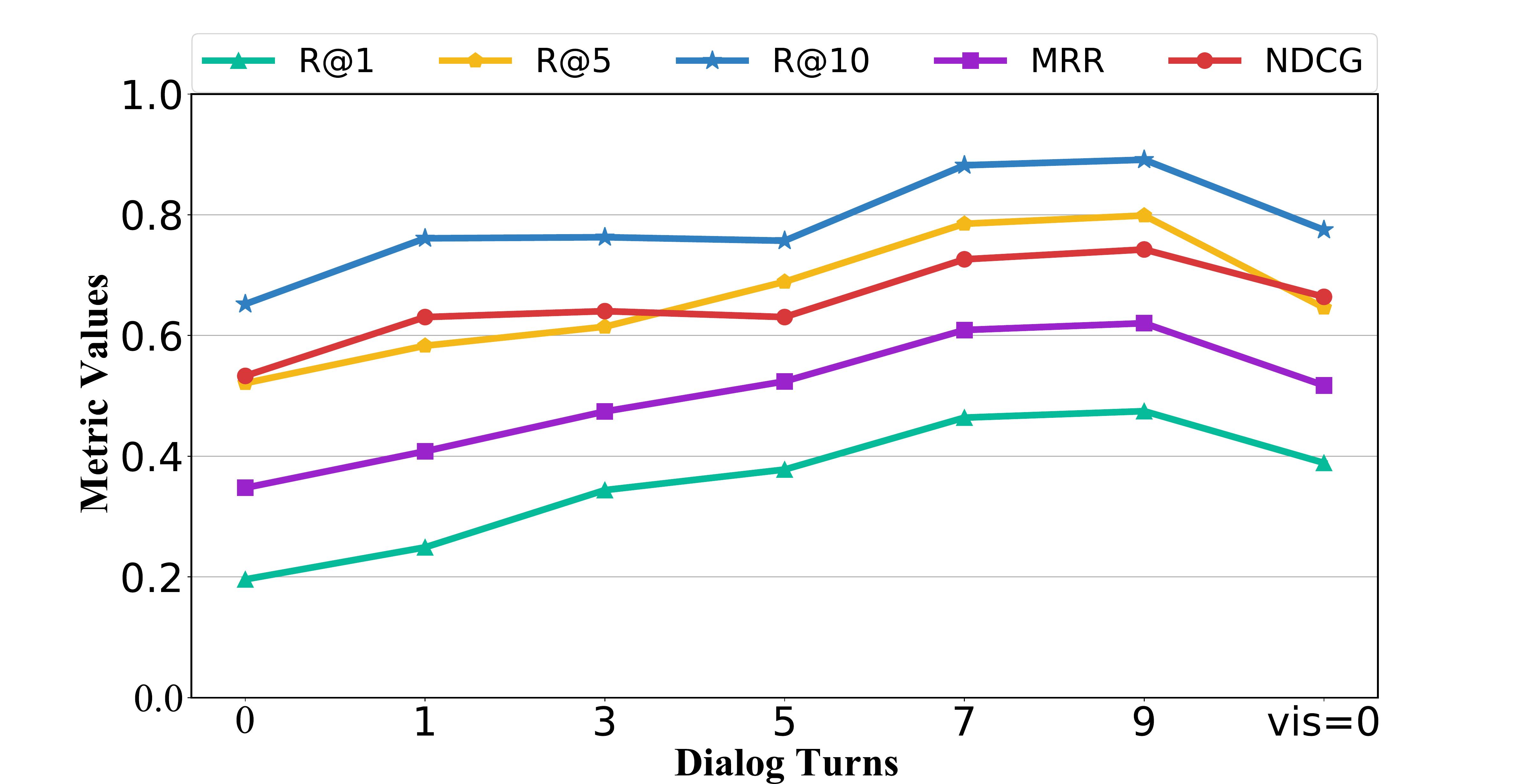}
 	\caption{Ablation study. $\text{vis}=0$ denotes the input includes 9 dialog turns with no image while others denote the different turns with images.}
	\label{fig2}

\end{figure}

In Fig \ref{fig2}, we represent how dialog history affects VD tasks. VU-BERT is trained with different dialog turns with image input and test in v1.0 validation set according to the dense annotation. We found that the metrics with only one turn or no dialog history is lower than other multi-turn dialog history. That is to say directly employing the models specific to the VQA and VC for VD tasks will not achieve better results. Also, more does not necessarily mean better for the dialog history. When $\text{turns}>7$, the metric values of Recall@K, MRR and NDCG gradually become flat, which may be caused by the limitation of input length. Besides, we compare the performance with the image input or not. As shown in Fig \ref{fig2}, the result of $\text{vis}=0$ is obviously lower when the dialog turns are both 9, which proves that VU-BERT takes advantage of multi-modal information to improve performance.

\begin{table}[]\centering
	\setlength{\tabcolsep}{1.5mm}{
		\caption{The parameters of transformer-based architecture}
		\begin{tabular}{llll}
			\hline
			Model     & Visdial-Bert & \multicolumn{1}{c}{VD-Bert} & VU-Bert \\
			\# Parameters & 315.26M  & 174.51M                 & \textbf{133.94M } \\ \hline
	\end{tabular}}	
	\label{tab:my-table4}
	
\end{table}

\section{conclusion}
\label{sec:foot}
In this paper, we present a simple visual dialog architecture VU-BERT with the effective fusion of two modalities. VU-BERT is comparable to competitors on VD tasks which are heavily equipped with convolution networks like Faster R-CNN. Besides, we explore the effect of dialog turns and found VQA and VC models are not suitable applied directly in VD tasks. This paper is an initial attempt to explore patch embedding adapted to VD tasks and we ask for future work on VD to focus more on the light model and deep interactions between images and dialog history with no convolution.

\section{acknowledgment}
This paper is supported by the Key Research and Development Program of Guangdong Province under grant No. 2021B0101400003 and the National Key Research and Development Program of China under grant No. 2018YFB0204 403. Corresponding author is Jianzong Wang from Ping An Technology (Shenzhen) Co., Ltd (jzwang@188.com).
%\vfill\pagebreak

\bibliographystyle{IEEEbib}
\bibliography{refs}

\end{document}